\documentclass[12pt]{article}
\usepackage[table,xcdraw]{xcolor} 
\usepackage{geometry}

\usepackage{booktabs}
\usepackage{array}
\usepackage{amsmath,amssymb}

\def\spacingset#1{\renewcommand{\baselinestretch}{#1}\small\normalsize}
\usepackage{amsthm}
\usepackage{rotating} 
\usepackage{subcaption} 
\usepackage{amsmath}
\usepackage{graphicx,psfrag,epsf}
\usepackage{enumerate}
\usepackage{tikz}
\usepackage{algorithm}
\usepackage{algpseudocode}
\usetikzlibrary{shapes.geometric, arrows.meta, positioning, shadows, backgrounds, fit,calc}
\usetikzlibrary{trees}

\tikzset{
    decision/.style = {diamond, draw, fill=blue!20, text width=3.5em, text centered, inner sep=1pt, drop shadow, font=\footnotesize},
    block/.style = {rectangle, draw, fill=green!20, text width=4em, text centered, rounded corners, minimum height=2.2em, drop shadow, font=\footnotesize},
    cloud/.style = {ellipse, draw, fill=orange!20, minimum height=1.6em, minimum width=4.5em, drop shadow, font=\footnotesize},
    line/.style = {draw, -stealth, thick, line width=0.7pt, shorten >=2pt, shorten <=2pt},
    desc/.style = {above, sloped, text width=4em, align=center, font=\tiny}
}
\usepackage{caption}
\usepackage{amssymb}
\usepackage{natbib}
\usepackage{color}
\usepackage{mathrsfs}  
\usepackage{stmaryrd}

\usepackage{bbold}
\usepackage{bbm}
\usepackage{multirow}
\usepackage{hyperref}

\newcommand{\I}{\mathbb{I}}
\newcommand{\pr}{\Pr}
\newcommand{\Err}{\mathcal{E}}

 \usepackage[utf8]{inputenc}

\newtheorem{assumption}{Assumption} 
 
\newtheorem{theorem}{Theorem}
 
\newtheorem{prop}{Proposition} 
\newtheorem{remark}{Remark}
\newtheorem{corollary}{Corollary}



\def \X{\mathbf{X}}

\def \I{\mathbb{I}}

\def\RR{\mathbb{R}}

\def \EE {\mathbb{E}}

\def \HH {\mathscr{H}}

\def \DR {{\rm DR}}

\newcommand{\indep}{\perp\!\!\!\perp}

\graphicspath{{figures/}}

\begin{document}
\spacingset{1}

\title{Cost-optimal Sequential Testing via Doubly Robust Q-learning}

\author{
   Doudou Zhou\textsuperscript{1}, Yiran Zhang\textsuperscript{1}, 
   Dian Jin\textsuperscript{1}, Yingye Zheng\textsuperscript{2}, Lu Tian\textsuperscript{3}\footnotemark[1],
    Tianxi Cai\textsuperscript{4}\thanks{Corresponding authors: \href{mailto:lutian@stanford.edu}{lutian@stanford.edu}; \href{mailto:tcai@hsph.harvard.edu}{tcai@hsph.harvard.edu}}  \\
    \small 
    \textsuperscript{1} Department of Statistics and Data Science, National University of Singapore  \\
    \small 
    \textsuperscript{2} Public Health Sciences Division, Fred Hutch \\
    \small 
    \textsuperscript{3} Department of Biomedical Data Science, Stanford University School of Medicine\\
      \small
    \textsuperscript{4} Department of Biostatistics, Harvard T.H. Chan School of Public Health}

\date{}

\maketitle
\bigskip
\begin{abstract}
Clinical decision-making often involves selecting tests that are costly, invasive, or time-consuming, motivating individualized, sequential strategies for what to measure and when to stop ascertaining. We study the problem of learning cost-optimal sequential decision policies from retrospective data, where test availability depends on prior results, inducing informative missingness. Under a sequential missing-at-random mechanism, we develop a doubly robust Q-learning framework for estimating optimal policies. The method introduces path-specific inverse probability weights that account for heterogeneous test trajectories and satisfy a normalization property conditional on the observed history. By combining these weights with auxiliary contrast models, we construct orthogonal pseudo-outcomes that enable unbiased policy learning when either the acquisition model or the contrast model is correctly specified. We establish oracle inequalities for the stage-wise contrast estimators, along with convergence rates, regret bounds, and misclassification rates for the learned policy. Simulations demonstrate improved cost-adjusted performance over weighted and complete-case baselines, and an application to a prostate cancer cohort study illustrates how the method reduces testing cost without compromising predictive accuracy.
\end{abstract}


\spacingset{1.45} 

\newpage

\section{Introduction}

Accurate prediction in clinical medicine often necessitates a battery of measurements that vary widely in cost, invasiveness, and availability. While the goal of diagnostic medicine is to reduce uncertainty, the modern healthcare landscape is increasingly defined by the tension between predictive precision and the burgeoning costs of care. In the US, healthcare spending per capita remains the highest in the word, yet clinical outcomes, such as life expectancy and the management of chronic conditions, do not consistently outperform those of European counterparts who utilize more conservative diagnostic protocols \citep{papanicolas2018health}. 

Evidence suggests that the US healthcare system is characterized by ``diagnostic intensity,'' ordering significantly more laboratory tests and specialized imaging per patient encounter than systems in the United Kingdom or Germany \citep{anderson2019it}. This disparity persists even when outcomes are adjusted for disease severity, suggesting that a substantial portion of medical testing may be redundant or of low marginal value \citep{smart2025inappropriate,blumenthal2024portrait}. Furthermore, many diagnostic procedures are invasive and carry non-negligible risks. Prostate cancer (PCa) diagnosis serves as a poignant motivating example. Traditionally, patients with elevated prostate-specific antigen (PSA) levels are referred for a transrectal ultrasound-guided biopsy, an invasive procedure associated with risks of infection, bleeding, and significant patient morbidity \citep{loeb2011complications}. Because the prevalence of clinically significant PCa is relatively low in general screened populations, a substantial proportion of these biopsies yield no actionable findings, representing an unnecessary clinical burden
\citep{anderson2019it}.

To mitigate these inefficiencies, current clinical guidelines from the National Comprehensive Cancer Network (NCCN) recommend a sequential, risk-stratified approach. This involves the use of secondary blood- and urine-based biomarkers or multiparametric magnetic resonance imaging (mpMRI) to determine if a biopsy is truly warranted \citep{carroll2016nccn}. However, as more tests become available, each with distinct accuracy profiles, costs, and levels of invasiveness, practitioners face an individualized decision problem: given the results of inexpensive or low-risk tests already obtained, should one stop and predict risk, or acquire a more expensive test?

Traditional statistical methods for cost reduction primarily rely on static feature selection, such as stepwise procedures or regularization methods via lasso \citep{tibshirani1996regression}. While these methods successfully yield parsimonious models, they are fundamentally limited by their ``one-size-fit-all'' nature, fixing the same subset of predictors for all individuals and disregarding the unique, real-time information revealed for a specific subject. To introduce a temporal element, sequential extensions based on pre-specified test orders \citep{lloyd2019use, grundy20192018} have been developed to allow for staged evaluation; however, these remain rigid and non-adaptive, as they cannot pivot based on intermediate results. In contrast, tree-based methods such as CART \citep{breiman2017classification} offer a degree of adaptivity by branching on observed variables, yet their predictive performance in complex clinical tasks is often inadequate compared to recent alternatives \citep{weng2017can}. 

Recent work in machine learning has sought to explicitly model prediction under budget constraints. Approaches such as cost-aware feature selection and active feature-value querying \citep{xu2012greedy, huang2018active, kachuee2018dynamic} aim to balance accuracy against the cost of additional measurements. These methods, however, typically assume either fully observed training data or an interactive environment in which unmeasured features can be queried on demand. In contrast, data from clinical applications are collected retrospectively, where the availability of specific tests often depends on the results of prior measurements. This leads to data that are \emph{informatively missing}: the presence of a test record itself carries information about patient risk and prior clinical decision-making. Standard methods that treat missingness as independent of unobserved values (e.g. missing-at-random) can therefore introduce significant bias.

 Sequential decision-making optimization is often formalized through the broader lens of reinforcement learning (RL) \citep{szepesvari2022algorithms,sutton1998introduction}. Dynamic Treatment Regime (DTR) methodology represents a prominent special case within this framework, focused on optimizing multistage interventions \citep{murphy2003optimal, zhang2013robust, zhao2015new}.  While DTR methods such as Q-learning and its refinements \citep{ye2024stage} effectively manage evolving patient histories, they are primarily designed for treatment selection, actions intended to alter patient state, rather than the acquisition of information. Consequently, the problem of sequential test acquisition requires a distinct formulation: the goal is not to intervene, but to reduce uncertainty. Closely related work in adaptive prediction \citep{cai2025dynamic} has applied RL strategies to this specific task of choosing tests sequentially. However, these analyses generally assume complete data availability, implying that outcomes under any hypothetical test sequence can be evaluated directly. 
 These approaches therefore do not address the central challenge of retrospective diagnostic data, where unacquired tests are genuinely missing because clinicians determined they were not indicated. Identifying optimal testing policies in such settings requires modeling the acquisition mechanism and correcting for history-dependent, informative missingness, which is the focus of this work.

This perspective suggests formulating the data-collection mechanism as a coarsening-at-random (CAR) or sequential missing-at-random (sequential MAR) process, in which each acquisition decision depends only on the observed history. Under such conditions, unbiased learning remains possible by correcting for the acquisition mechanism. Classical work on inverse probability weighting (IPW) and augmented estimators \citep{robins2000msm, bang2005dr} provides identification tools for longitudinal and coarsened data, while recent developments in doubly robust and orthogonal machine learning \citep{ChernozhukovDML2018, kennedy2023ejs} give general strategies for constructing pseudo-outcomes that are robust to nuisance misspecification. These results, however, do not directly yield the stage-wise value contrasts that govern optimal test acquisition, nor do they address the multiple acquisition paths that lead to the same observed feature set in sequential diagnostic workflows.

In this paper we develop \textbf{COST-Q} (\textbf{C}ost-\textbf{O}ptimal \textbf{S}equential \textbf{T}esting via Doubly Robust \textbf{Q}-learning), integrating semiparametric identification with dynamic programming. We formalize sequential test acquisition as a dynamic decision problem in which the key statistical objects are stage-specific contrast functions, defined as conditional expectations of cost-augmented loss differences that determine whether an additional test is worthwhile. Identification under sequential MAR is achieved via path-aware propensity weights that are computable from observed histories and normalized so that they aggregate multiple acquisition routes to a given feature set. These weights are combined with auxiliary contrast models to construct orthogonal, doubly robust pseudo-outcomes that remain unbiased when either the acquisition mechanism or the contrast model is correctly specified. A backward Q-learning procedure with cross-fitted nuisance estimation then yields consistent estimators of the stage-wise contrasts and an associated sequential decision rule. We establish double-robustness and convergence guarantees for the resulting estimators, and show in simulations and a real diagnostic application that the learned policies substantially reduce test costs without sacrificing predictive performance.

\section{Method}
\label{sec:method}

We now present COST-Q, a framework that formulates the sequential testing problem through stage-wise contrast functions estimated via path-specific importance weights and doubly robust pseudo-outcomes, optimized through a backward Q-learning procedure to learn the diagnostic policy.

\subsection{Problem setup}
\label{sec:problem}

We consider a sequential testing problem with a baseline feature block $X_0 \in \RR^{p_0}$, which is always observed at no cost ($c_0=0$), and $M$ optional diagnostic test blocks. For ease of presentation, we focus on $M=2$, but note that the methodology extends to $M>2$ by backward induction over ordered stages. Let $X_1 \in \RR^{p_1}$ and $X_2 \in \RR^{p_2}$ denote the two optimal test blocks. In clinical applications, $X_0$ may include demographics and routine laboratory values, while $X_1$ and $X_2$ represent specialized procedures. The acquisition process is defined by a sequence of decisions $(S_1, S_2)$, where $S_j\in \{0,1,2\}$ denotes the $j$th test selected, and $S_j=0$ indicates that the testing process has terminated. The valid decision space is constrained such that no test is repeated yielding the set of possible paths:
\[
\mathcal{P} = \{(0,0), (1,0), (2,0), (1,2), (2,1)\}.
\]
To streamline the analysis of history-dependent missingness, we map these acquisition paths to a discrete set of \textbf{information states} $s \in \mathcal{S}=\{0,1,2,12\}$. Each state $s$ defines the subset of observed features and the cumuative costs $C_s$, as summarized in Table~\ref{tab:info_states}. Figure S1 in Supplementary S.1 further details the acquisition paths and the induced patterns.

\begin{table}[h]
\centering
\caption{Information State ($s$), Acquisition Path ($(S_1,S_2)$), Observed Feature Vector ($\X_s$), and Cumulative Costs ($C_s$).}
\label{tab:info_states}
\begin{tabular}{@{}llll@{}}\hline
\textbf{State $s$} & \textbf{Path $(S_1, S_2)$} & \textbf{Observed Feature ($\mathbf{X}_s$)} & \textbf{Cumulative Cost ($C_s$)} \\ \hline
0 & $(0,0)$ & $\X_0 = X_0$ & $0$ \\
1 & $(1,0)$ & $\X_{1} = (X_0^\top, X_1^\top)^\top$ & $c_1$ \\
2 & $(2,0)$ & $\X_{2} = (X_0^\top, X_2^\top)^\top$ & $c_2$ \\
12 & $(1,2)$ or $(2,1)$ & $\X_{12} = (X_0^\top, X_1^\top, X_2^\top)^\top$ & $c_1 + c_2$ \\ \hline
\end{tabular}
\end{table}

\paragraph{Data collection and identification} A primary challenge in this setting is that the terminal state $s=12$ can be reached via two distinct acquisition paths. While the final feature vector $\mathbf{X}_{12}$ is identical, the intermediate information used to justify the second test—and thus the underlying selection mechanism—differs by path. Because clinicians order tests adaptively, the observed data $\mathcal{D} = \{(Y^i, \mathbf{X}_{s_i}^i, S_1^i, S_2^i)\}_{i=1}^n$ exhibit informative missingness. To formalize this mechanism, we define state-wise propensity scores: 
\begin{equation*}
\begin{aligned}
    \pi^*_{j\mid 0}(\mathbf{x}_0) &= \pr(S_1=j \mid \mathbf{X}_0=\mathbf{x}_0), \quad j \in \{0, 1,2\},  \\
    \pi^*_{j^c\mid j}(\mathbf{x}_j) &= \pr(S_2=j^c \mid S_1=j, \mathbf{X}_j=\mathbf{x}_j), \quad j\in \{1, 2\}, j^c \in \{1, 2\}\setminus\{j\}. 
\end{aligned}    
\end{equation*}
Unbiased identification of the optimal policy from retrospective data requires that acquisition decisions depend only on the observed history, a sequential version of the Missing at Random (MAR) assumption. 
\begin{assumption}[Sequential MAR]
\label{assump:id}
The acquisition variables depend only on previously observed history: 
\[
S_1 \indep (X_1,X_2,Y)\mid \mathbf{X}_0,\quad
S_2 \indep (X_{j^c},Y)\mid (\mathbf{X}_j,S_1=j) \quad \text{for distinct  }  j, j^c\in\{1,2\}.
\]
\end{assumption}
In clinical practice, this implies physicians order tests based solely on available measurements rather than unobserved patient characteristics.

\subsection{Optimal policy and contrast functions}

At each state $s\in \mathcal{S}$, we derive a predictive model for $m_s(\mathbf{x}_s)=\EE[Y \mid \mathbf{X}_s=\mathbf{x}_s]$ to predict the outcome. Let $L(Y,m(\mathbf{X}))$ denote a standard loss (e.g. squared error or cross-entropy). We define the cost-augmented loss at state $s$ as
\begin{equation}
\label{def:error}
\Err_s(Y,\mathbf{X}_s) = L(Y,m_s(\mathbf{X}_s)) + C_s.
\end{equation}
We assume $\mathbb{E}[\Err_s^2]<\infty$ and treat $C_s$ as pre-calibrated to the scale of $L$. Our objective is to learn an optimal prescriptive policy $\mathbf{d} = (d_0, d_1, d_2)$, with $d_j$ a function of observed $\X_j$, that minimizes the expected cost-augmented loss by selecting actions from the the admissible set 
$$
\mathcal{A}(s) = \{0\} \cup (\{1, 2\} \setminus \{s\}),
$$ 
at each non-terminal state $s$. The optimal policy is governed by the Bellman principle. 

Let $Q^\mathbf{d}_s(\X_s)$ denote the expected cost-augmented loss when starting from state $s$ and following policy $\mathbf{d}$ given observed $\X_s$. In the second stage, the process is already in state $s=j \in \{1,2\}$, meaning test $j$ has been acquired. The decision rule $d_j$ then selects between two possible actions: terminating the process ($d_j=0$) or acquiring the remaining test block ($d_j=j^c$, where $j^c = \{1,2\} \setminus \{j\}$). The second-stage value function is thus defined as:
\begin{align*}
Q_{j}^{\mathbf{d}}(\mathbf{X}_j)
&= \mathbb{E}\left[ 
\Err_{j} \mathbb{I}\left(d_{j} =0\right)
+ \Err_{12} \mathbb{I}\left(d_{j} =j^c\right) \mid \mathbf{X}_j \right],
\end{align*}
This formulation shows that $Q_j^{\mathbf{d}}$ is a switch between the current expected error and the expected error after full data acquisition, depending on the policy's decision at that state.

By the Bellman  principle, the optimal second-stage actions 
\begin{equation*}
d_j^*(\X_j)
= j^c\times\mathbb{I}\left( \Delta_{j^c \mid j}^*(\X_j) <0 \right), 
\end{equation*}
driven by the \emph{second-stage contrast function}
\begin{equation*}
\Delta_{j^c\mid j}^*(\X_j) := \mathbb{E}\left( T_{j^c \mid j } \mid \mathbf{X}_j\right) \; \text{ with } T_{j^c \mid j } = \Err_{12}-\Err_j .
\end{equation*} 
The corresponding optimal continuation values $Q_j^*(\mathbf{x}_j)$ is then 
\begin{equation*}
Q_j^*(\mathbf{x}_j) = \mathbb{E}\!\left[
\Err_{j}\,\mathbb{I}(d_{j}^* =0)
+ \Err_{12}\,\mathbb{I}(d_{j}^* =j^c)
\;\mid\; \mathbf{X}_j=\mathbf{x}_j \right]  
=\mathbb{E}\left(\Err_{j}\mid \mathbf{X}_j=\mathbf{x}_j\right)\wedge \mathbb{E}\left(\Err_{12}\mid \mathbf{X}_j=\mathbf{x}_j \right),
\end{equation*}
where $a \wedge b := \min(a, b)$. At $s=0$, the value function incorporates these continuation values:
\begin{align*}
Q_{0}^{\mathbf{d}}(\X_0)
=& \mathbb{E}\!\left[
\Err_{0}\,\mathbb{I}\left(d_{0}=0 \right)
+ Q^{\mathbf{d}}_{1}\left(\mathbf{X}_1)\,\mathbb{I}(d_{0} =1 \right)
+ Q^{\mathbf{d}}_{2}\left(\mathbf{X}_2)\,\mathbb{I}(d_{0} =2 \right)
\;\mid\; \mathbf{X}_0 \right].
\end{align*}
The optimal first-stage decision $d_0^*(\mathbf{x}_0)$ is 
\[
d_0^*(\mathbf{x}_0)
=\begin{cases}
j,  & \text{if }\Delta_{j\mid 0}^*(\mathbf{x}_0) < \Delta_{j^c\mid 0}^*(\mathbf{x}_0)\wedge 0, \quad j\in\{1,2\}, j^c \in \{1, 2\}\setminus\{j\}, \\[2pt]
0, & \text{if } 0<\Delta_{1\mid 0}^*(\mathbf{x}_0) \wedge \Delta_{2\mid 0}^*(\mathbf{x}_0) ,
\end{cases}
\]
determined by the first-stage contrast functions: 
\begin{equation*}
\Delta_{j\mid 0}^*(\mathbf{x}_0) 
:= \mathbb{E}\big( T_{j \mid 0} \,\mid\, \mathbf{X}_0=\mathbf{x}_0\big) \; \text{ with }  T_{j \mid 0}  = Q_j^*(\mathbf{X}_j)-\Err_0, 
\quad j\in\{1,2\}.
\end{equation*}
Thus the optimal policy $\mathbf{d}^*=(d_0^*,d_1^*,d_2^*)$ is fully characterized by the contrast functions $\Delta_{j^c\mid j}^*(\cdot)$ and $\Delta_{j\mid 0}^*(\cdot)$, which are the primary targets to be estimated.

\subsection{COST-Q estimation strategy} 

\subsubsection{From idealized contrasts to doubly robust estimation}
\label{sec:ideal_to_dr}
To motivate our estimators, we first consider an idealized setting where the complete feature vector $\X_{12}$ is observed for all subjects. In this complete-data case \citep{cai2025dynamic}, the optimal decision at state $s=1$ is determined by the contrast $\Delta_{2\mid1}^*(\mathbf{x}_1) = \EE\big(\Err_{12}-\Err_1 \,\mid\, \mathbf{X}_1=\mathbf{x}_1\big)$. With full observability, one would simply form the labels $T_{2\mid1}^i := \Err_{12}^i - \Err_1^i$ and estimate $\Delta_{2\mid1}^*(\cdot)$ via standard regression on $\mathbf{X}_1^i$.
In retrospective clinical settings, however, these ideal labels are only partially observed. Specifically, $T_{2\mid1}$ is available only for units with $s=12$. A naive regression on this subpopulation targets $\EE\big(T_{2\mid1}\mid \mathbf{X}_1, s=12\big)$ rather than the marginal expectation, introducing selection bias. Under Assumption~\ref{assump:id}, we correct this via a doubly robust framework using path-specific importance weights.
We construct pseudo-outcomes of the form:
\begin{equation}
\label{eq:po-generic}
\Phi(\eta) = \Delta(Z) + w(V;\pi)\big[T - \Delta(Z)\big],
\end{equation}
where $Z$ represents the conditioning covariates for the target contrast $\Delta^*(z)=\EE(T\mid Z=z)$, $T$ is the partially observed label, and $w(V;\pi)$ is a stage- and path-specific weight. 

\subsubsection{Identification and construction of the importance weights}
\label{subsec:weights}
Unbiased estimation requires identifying conditional means from history-dependent missing data. The following result guides our sample construction for both the core prediction models and the contrast regressions.

\begin{prop}[Conditional-mean invariance under sequential MAR]
\label{thm:ignorable_invariance_all}
Under Assumption~\ref{assump:id}, the following hold almost surely for $j\in\{1,2\}$:
\begin{equation*}
    \begin{aligned}
& \EE(Y \mid \mathbf X_j) = \EE(Y \mid \mathbf X_j, S_1=j) = \EE(Y \mid \mathbf X_j, S_1=j, S_2=j^c),\\
& \EE(T_{j^c\mid j} \mid \mathbf X_j) = \EE(T_{j^c\mid j} \mid \mathbf X_j, S_1=j) = \EE(T_{j^c\mid j} \mid \mathbf X_j, S_1=j, S_2=j^c), \\
& \EE(T_{j\mid0}\mid \mathbf X_0,S_1=j) = \EE(T_{j\mid0}\mid \mathbf X_0),
\\
& \EE(Y\mid \mathbf X_{12}, (S_1,S_2)=(1,2))
=
\EE(Y\mid \mathbf X_{12}, (S_1,S_2)=(2,1))
=
\EE(Y\mid \mathbf X_{12}).
\end{aligned}
\end{equation*}
\end{prop}
To adjust for informative acquisition, COST-Q relies on inverse probability weighting. Because a single feature set may be reached via multiple paths, weights are defined pathwise using the history available at each decision point:
\begin{alignat}{2}\label{eq:w12}
& \mbox{\bf Second stage weights}: &\quad& w_{j^c\mid j}^* = \frac{\mathbb{I}(S_2=j^c)}{\pi^*_{j^c\mid j}(\mathbf{X}_j)} \\
\label{eq:w1}
& \mbox{\bf First stage weights}: &\quad& w_{j\mid 0}^* = \frac{\mathbb{I}(S_1=j)}{\pi^*_{j\mid 0}(\mathbf{X}_0)}
\end{alignat}
Defining the doubly robust pseudo-outcomes $\widehat{\Phi}_{j^c\mid j}$ and $\widehat{\Phi}_{j\mid 0}$ using these weights ensures they are unbiased for their respective target contrasts whenever either the propensity or contrast model is correctly specified (Proposition~\ref{prop:dr_property}).

\subsubsection{Estimating the optimal policy $\mathbf{d}^*(\cdot)$ }
\label{subsec:d}

We estimate the optimal decision rule via a backward Q-learning scheme. Starting from the final decision stage and moving backward, we instantiate the generic pseudo-outcome \eqref{eq:po-generic} with suitable $(Z,T,w)$ at each stage and then regress $\Phi$ on $Z$ to obtain estimators of the corresponding contrast functions.
To ensure the COST-Q estimators are both efficient and unbiased, we employ a $K$-fold cross-fitting procedure that decouples the nuisance model estimation from the construction of pseudo-outcomes \citep{ChernozhukovDML2018}. We partition the data $\mathcal{D}$ into $K$ disjoint folds $\{\mathcal{D}_k\}_{k=1}^K$ and, for each fold $k$, train the nuisance functions $\eta^{(-k)} = (\widehat{\pi}^{(-k)}, \widehat{\Delta}^{(-k)})$ using only the complementary data $\mathcal{D} \setminus \mathcal{D}_k$. The specific training samples for these nuisance models are defined as follows:
\begin{itemize}
\item {\bf Propensity models ($\widehat{\pi}$):} The second-stage propensity $\widehat{\pi}_{j^c\mid j}^{(-k)}(\mathbf{x}_j)$ is fitted using all subjects in $\mathcal{D} \setminus \mathcal{D}_k$ with $S_1=j$, while the first-stage propensity $\widehat{\pi}_{j\mid 0}^{(-k)}(\mathbf{x}_0)$ is fitted using $\mathcal{D} \setminus \mathcal{D}_k$.
\item {\bf Contrast models ($\widehat{\Delta}$):} The second-stage contrast $\widehat{\Delta}_{j^c\mid j}^{(-k)}(\mathbf{x}_j)$ is fitted using only subjects in $\mathcal{D} \setminus \mathcal{D}_k$ who followed the ordered path $(S_1 = j, S_2 = j^c)$. Analogously, the first-stage contrast $\widehat{\Delta}_{j\mid 0}^{(-k)}(\mathbf{x}_0)$ is fitted using subjects with $S_1=j$.
\end{itemize}
This out-of-fold estimation ensures that for each subject $i \in \mathcal{D}_k$, the pseudo-outcomes $\widehat{\Phi}^i$ are computed using nuisance parameters $\widehat{\eta}^{(-k)}$ that are independent of $(Y^i, S_1^i, S_2^i)$. By cycling through all $K$ folds, we obtain a full set of doubly robust pseudo-outcomes $\{\widehat{\Phi}^i\}_{i=1}^n$ that preserve the orthogonality required for valid downstream inference and policy learning.

\paragraph{Step 1: Estimation of second stage rule $d_1^*(\cdot)$ and $d_2^*(\cdot)$}

We begin with the second-stage contrasts 
$\Delta_{j^c\mid j}^*(\mathbf{x}_j) = \mathbb{E}[T_{j^c\mid j} \mid \mathbf{X}_j=\mathbf{x}_j]$, 
whose label $T_{j^c\mid j}=\Err_{12}-\Err_j$ is observed for subjects on the ordered path $(S_1,S_2)=(j,j^c)$. For each $i$th subject in $\mathcal{D}_k$ and each distinct pair $(j,l)$, define the doubly robust pseudo-outcome
\begin{equation}
\label{eq:phi_lj}
\widehat{\Phi}_{j^c\mid j}^{i}
=\widehat{\Delta}_{j^c\mid j}^{(-k)}(\mathbf{X}_j^i)
+\widehat w_{j^c\mid j}^{\,i,(-k)}\left[\left(\Err_{12}^i-\Err_j^i \right)-\widehat{\Delta}_{j^c\mid j}^{(-k)}(\mathbf{X}_j^i)\right], \quad
\mbox{with 
$w_{j^c\mid j}^{\,i,(-k)}=
\frac{\mathbb{I}(S_2^i=j^c)}
     {\widehat{\pi}^{(-k)}_{j^c\mid j}(\mathbf{X}_j^i)}$.
}
\end{equation}
We estimate the contrast functions by empirical risk minimization.  
Let $\HH_j^d$ denote a function class and let $\ell$ be a generic loss.  
With out-of-fold pseudo-outcomes $\{\widehat{\Phi}_{j^c\mid j}^i\}_{i=1}^n$ calculated for subjects with $S_1=j$, the doubly robust estimator solves
\begin{equation}
\widehat{\Delta}_{\DR,j^c\mid j}
    = \arg\min_{\Delta\in\HH_j^d}
      \sum_{i=1}^n \ell\!\left(\widehat{\Phi}_{j^c\mid j}^i,\; \Delta(\mathbf{X}_j^i)\right)\I(S_1^i = j).
\label{eq:delta_dr_lj}
\end{equation}
The corresponding second-stage rule is
\begin{equation}
\label{eq:dj_dr}
\widehat d_j(\mathbf{x}_j)
    = j^c\times \mathbb{I}\!\left(\widehat{\Delta}_{\DR,j^c\mid j}(\mathbf{x}_j)<0\right),     \qquad j \in\{1,2\}, j^c \in \{1, 2\}\setminus\{j\}.
\end{equation}
The choices of $\HH_j^d$ and $\ell$ depend on the structure of the pseudo-outcomes and the desired regularity properties of the estimator.

For each outer fold $k$, we additionally define the fold-specific second-stage contrast estimator
$\widehat{\Delta}_{DR,j^c\mid j}^{(-k)}$ by repeating Step 1 using only the training sample
$\mathcal{D}\setminus\mathcal{D}_k$ (with cross-fitting carried out internally within that sample if desired).
The corresponding fold-specific second-stage rule is
\[
\widehat d_j^{(-k)}(\mathbf{x}_j)
=
j^c \times \mathbb{I}\!\left(\widehat{\Delta}_{\DR,j^c\mid j}^{(-k)}(\mathbf{x}_j)<0\right),
\qquad j\in\{1,2\}.
\]
This definition ensures that $\widehat d_j^{(-k)}$ is constructed independently of all observations in
$\mathcal{D}_k$.

\paragraph{Step 2: Doubly robust estimation of $d_0^*(\cdot)$}
The initial rule depends on the contrast
$\Delta_{j\mid0}^*(\mathbf{x}_0)=\EE\left(Q_j^*-\Err_0\mid \mathbf{X}_0=\mathbf{x}_0\right)$,
where $Q_j^*$ is the latent optimal future value.
To preserve sample-splitting, for each fold $k$ we first construct the fold-specific
second-stage rule $\widehat d_j^{(-k)}$ using only the training sample
$\mathcal{D}\setminus\mathcal{D}_k$, as defined above.
Then, for each subject $i\in\mathcal{D}_k$, the estimated continuation value is
\begin{align}
\widetilde{Q}_{j}^{\,i,(-k)}
= &\Err_{j}^i \cdot \mathbb{I}\left(\widehat{d}_{j}^{(-k)}(\mathbf{X}_j^i)=0\right)
   + \left[\Err_{j}^i+\widehat{\Delta}_{\DR,j^c\mid j}^{(-k)}(\mathbf{X}_j^i)\right]\cdot
     \mathbb{I}\left(\widehat{d}_{j}^{(-k)}(\mathbf{X}_j^i)=j^{c}\right) \nonumber\\
= &\Err_{j}^i
   + \widehat{\Delta}_{\DR,j^c\mid j}^{(-k)}(\mathbf{X}_j^i)\cdot
     \mathbb{I}\left(\widehat{d}_{j}^{(-k)}(\mathbf{X}_j^i)=j^{c}\right).
\label{eq:Qtilde}
\end{align}
Note that $\widetilde{Q}_{j}^{\,i,(-k)}$ is undefined if $\mathbf{X}_j^i$ is unobserved.
We construct the estimated nuisance propensity function $\widehat{\pi}_{j\mid 0}^{(-k)}(\cdot)$ by regressing $\mathbb{I}(S_1^i=j)$ on $\mathbf{X}_0^i$ using all subjects in $\mathcal{D}\setminus\mathcal{D}_k$. Similarly, we obtain the estimated nuisance function $\widehat{\Delta}_{j\mid0}^{(-k)}(\cdot)$ by regressing $(\widetilde{Q}_{j}^{\,i,(-k)}-\Err_0^i)$ on $\mathbf{X}_0^i$ using all subjects in $\mathcal{D}\setminus\mathcal{D}_k$ for whom $S_1=j$ as detailed in Section~\ref{subsec:m}. 

With the nuisance models trained, we apply the same construction to $\mathcal{D}_k$. For the $i$th subject in  this fold, we compute $\widetilde{Q}_{j}^{\,i,(-k)}$ via \eqref{eq:Qtilde} and form the doubly robust pseudo-outcome:
\begin{equation}
\widehat{\Phi}_{j\mid0}^i = \widehat{\Delta}_{j\mid0}^{(-k)}(\mathbf{X}_0^i)
+\widehat w_{j\mid0}^{\,i,(-k)}\Big[(\widetilde{Q}_{j}^{\,i,(-k)}-\Err_0^i)-\widehat{\Delta}_{j\mid0}^{(-k)}(\mathbf{X}_0^i)\Big],  \quad
\mbox{with  
$\widehat w_{j\mid0}^{\,i,(-k)}=
\frac{\mathbb{I}(S_1^i=j)}{\widehat{\pi}^{(-k)}_{j\mid 0}(\mathbf{X}_0^i)}$.}
\label{eq:phi_upq_j0}
\end{equation}
If $S_1^i \neq j$, then $\widehat w_{j\mid0}^{\,i,(-k)} = 0$, so the pseudo-outcome reduces to the model prediction $\widehat{\Delta}_{j\mid0}^{(-k)}(\mathbf{X}_0^i)$.

Finally, pooling the pseudo-outcomes $\{\Phi_{j\mid0}^i\}_{i=1}^n$ across all folds, the first-stage contrasts are estimated by:
\begin{equation}
\widehat{\Delta}_{\DR,j\mid0}=\arg\min_{\Delta\in\HH_0^d}\sum_{i=1}^n \ell\Big(\widehat{\Phi}_{j\mid0}^i, \Delta(\mathbf{X}_0^i)\Big),\qquad j\in\{1,2\}.
\label{eq:delta_dr_j0}
\end{equation}
The first-stage rule is then
\begin{equation}
\label{eq:d0_dr}
\widehat d_0(\mathbf{x}_0)=
\begin{cases}
  j, & \text{if } \widehat{\Delta}_{\DR,j\mid0}(\mathbf{x}_0) < \widehat{\Delta}_{\DR,j^c\mid0}(\mathbf{x}_0)\wedge 0, \qquad j \in\{1,2\}, j^c \in \{1, 2\}\setminus\{j\} \\
    0, & \text{if } 0<\widehat{\Delta}_{\DR,1\mid0}(\mathbf{x}_0)\wedge \widehat{\Delta}_{\DR,2\mid0}(\mathbf{x}_0). 
\end{cases}
\end{equation}

\subsubsection{Estimation of core models and propensities}
\label{subsec:m}

If the core prediction models $m_s(\cdot)$ are not provided, we estimate them by unweighted empirical risk minimization (ERM) on the subset where the state-specific features are observed. Let $\mathcal I_s$ denote the corresponding index set. The estimator for each state $s\in\{0,1,2,12\}$ is
\begin{equation}
    \label{eq:m_unified}
    \widehat{m}_s
    =
    \arg\min_{m_s\in \mathcal{H}_s^m}
    \sum_{i \in \mathcal{I}_s} L\!\big(Y^i,\,m_s(\mathbf{X}_s^i)\big).
\end{equation}
Here $\mathcal I_0$ contains all subjects, while for $s\in\{1,2\}$ we take $\mathcal I_s=\{i:S_1^i=s\}$ and for $s=12$ we take $\mathcal I_{12}=\{i:s^i=12\}$. Proposition~\ref{thm:ignorable_invariance_all} implies that these sample choices target the desired conditional means: for $j\in\{1,2\}$, $\EE(Y\mid \mathbf X_j)=\EE(Y\mid \mathbf X_j,S_1=j)$, and  for the terminal state,
\[
\EE(Y\mid \mathbf X_{12},(S_1,S_2)=(1,2))
=
\EE(Y\mid \mathbf X_{12},(S_1,S_2)=(2,1))
=
\EE(Y\mid \mathbf X_{12}).
\]
Thus, $m_j$ can be fitted on subjects with $S_1=j$, while $m_{12}$ can be fitted by pooling both ordered terminal paths. Inverse probability weighting is not used at this stage, since informative acquisition is handled later through the doubly robust contrast estimation; incorporating weights here would mainly increase variance without clear benefit for downstream policy learning.

The same principle guides the contrast regressions. The second-stage contrast $\widehat{\Delta}_{j^c\mid j}$ is fitted using only the ordered-path subsample with $(S_1,S_2)=(j,j^c)$, which targets $\Delta_{j^c\mid j}^*(\cdot)=\EE(T_{j^c\mid j}\mid \mathbf X_j)$ from Proposition~\ref{thm:ignorable_invariance_all}. Pooling all subjects with $s=12$ would mix the reverse path $(S_1,S_2)=(j^c,j)$ and generally alter the conditional target, so we do not use pooled terminal-state regression for estimating $\Delta_{j^c\mid j}^*$.

Similarly, the first-stage contrast regression targets
$\Delta_{j\mid0}^*(\mathbf x_0)=\EE(T_{j\mid0}\mid \mathbf X_0=\mathbf x_0)$. This conditional mean is unchanged after conditioning on $S_1=j$, so the regression can be carried out on the subsample with $S_1=j$. In implementation, the oracle target $T_{j\mid0}$ is replaced by $\widetilde Q_j-\Err_0$.

Finally, the propensity scores entering the importance weights are estimated by cross-entropy ERM within the same $K$-fold cross-fitting scheme. For example, the second-stage propensity $\widehat{\pi}_{j^c \mid j}^{(-k)}(\mathbf{x}_1)$ is estimated on $\mathcal{D}\setminus\mathcal{D}_k$ by
\[
\widehat{\pi}_{j^c \mid j}^{(-k)}
=
\arg\max_{\pi \in \mathcal{H}^{\pi}_{j}}
\sum_{i \in \mathcal{D}\setminus\mathcal{D}_k,\; S_1^i=j}
\left\{
\mathbb{I}(S_2^i=0)\log\!\left[1-\pi(\mathbf{X}_1^i)\right]
+
\mathbb{I}(S_2^i=j^c)\log\!\left[\pi(\mathbf{X}_1^i)\right]
\right\}.
\]
The classes $\mathcal{H}^{\pi}_{j}$ can be flexibly specified, for example using neural networks with sigmoid or softmax output layers to enforce valid probability predictions. The complete COST-Q procedure is summarized in Algorithm S1 in Supplementary S.1.

\section{Theoretical Analysis}
\label{sec:theory}

We now establish the theoretical properties of COST--Q. The analysis has two goals. First, we derive oracle inequalities for the stage-wise doubly robust contrast estimators. Second, we translate these estimation error bounds into regret and policy misclassification guarantees for the learned sequential testing policy.

The results build on the identification setup introduced in Section~\ref{sec:method}, in particular Assumption~\ref{assump:id}, together with the following positivity condition.

\begin{assumption}[Positivity]
\label{assump:positivity}
There exists $\underline c\in(0,\tfrac12)$ such that, almost surely,
\[
\pi^*_{0\mid0}(\mathbf{X}_0),\ 
\pi^*_{1\mid0}(\mathbf{X}_0),\ 
\pi^*_{2\mid0}(\mathbf{X}_0),\ 
\pi^*_{2\mid1}(\mathbf{X}_1),\ 
\pi^*_{1\mid2}(\mathbf{X}_2)
\in [\underline c,1-\underline c].
\]
\end{assumption}

Assumption~\ref{assump:positivity} ensures that all admissible actions occur with nonvanishing conditional probability, so the inverse-probability weights are well defined and uniformly controlled.

\begin{remark}
To keep the main text concise, the formal operator-level definitions of the oracle benchmark and stable learner, together with the full technical assumptions, are deferred to Supplementary S.2. Beyond Assumptions~\ref{assump:id} and \ref{assump:positivity}, the oracle inequalities below rely on three high-level ingredients: 1) a stability condition on the final regression learner, 2) rate conditions on the cross-fitted nuisance estimators, and 3) a boundedness condition on the learner as an operator. Together, these ensure that perturbations from estimated pseudo-outcomes are controlled, yield double-robust consistency when at least one nuisance component is consistently estimated, and keep second-order bias manageable. The policy misclassification results further invoke Tsybakov-type margin conditions for the stage-wise contrasts and the first-stage pairwise gap. Formal statements are given in Definitions 1 and 2 and Assumptions 3--7 of Supplementary S.2. Specifically, Proposition~\ref{prop:dr_property} uses only Assumptions~\ref{assump:id} and \ref{assump:positivity}; Theorems~\ref{thm:oracle-stage2} and \ref{thm:oracle-stage1} further require the stability, nuisance-rate, and boundedness conditions; and Corollaries~\ref{thm:margin_rates_single} and \ref{coro:margin_costq} additionally rely on the margin assumptions.
\end{remark}

For a generic stage, let $Z$ denote the conditioning covariates and define the target regression function $\Delta^*(z):=\EE(T\mid Z=z)$. Let $\Phi^*$ denote the oracle pseudo-outcome constructed from the true nuisance functions, and let $\tilde\Delta:=\mathcal L_n\big(\{(\Phi_i^*,Z_i)\}_{i=1}^n\big)$ be the corresponding infeasible oracle estimator. Its oracle risk is
\[
R_n^*:=\Big\{\EE\big[\|\tilde\Delta-\Delta^*\|_{L_2(P)}^2\big]\Big\}^{1/2},
\]
where the expectation is taken with respect to the randomness of $\{(\Phi_i^*,Z_i)\}_{i=1}^n$ and any internal randomness of $\mathcal L_n$. Throughout, $\|f\|_{L_q(P)}$ denotes the $L_q$ norm with respect to the marginal distribution of $Z$, i.e. $\|f\|_{L_2(P)}^2:=\EE\{f(Z)^2\}$. We also denote the nuisance estimation errors by $r_\pi:=\max_k\|\widehat\pi_k-\pi_k^*\|_{L_2(P)}$ and $r_\Delta:=\|\widehat\Delta-\Delta^*\|_{L_2(P)}$. When needed, we index oracle risks by stage, writing $R_{n,j^c\mid j}^*$ and $R_{n,j\mid0}^*$.

In COST--Q, the second-stage problem uses $Z=\mathbf{X}_j$, $T=\Err_{12}-\Err_j$, and weight $w_{j^c\mid j}^*$ defined in \eqref{eq:w12}. The corresponding oracle pseudo-outcome is $\Phi_{j^c\mid j}^*  := \Phi_{j^c\mid j}(\Delta_{j^c\mid j}^*,  w_{j^c\mid j}^*)$ where 
\begin{align}\label{eq:phi_lj_o}
    \Phi_{j^c\mid j}(\Delta_{j^c\mid j}, w_{j^c\mid j}) =\Delta_{j^c\mid j}(\mathbf{X}_j)+ w_{j^c\mid j}\big[(\Err_{12}-\Err_j)-\Delta_{j^c\mid j}(\mathbf{X}_j)\big].
\end{align}

For the first-stage, we take $Z=\mathbf{X}_0$, $T=\widetilde Q_j^*-\Err_0$, and weight $w_{j\mid0}^*$ defined in \eqref{eq:w1}, where
\begin{equation}
\widetilde Q_j^*:=\Err_j + 0\wedge\Delta_{j^c\mid j}^*(\mathbf{X}_j)
\label{eq:oracle_upq}
\end{equation}
is the oracle continuation value at state $j$. The corresponding oracle pseudo-outcome is $\Phi_{j\mid0}^*:= \Phi_{j\mid0}(\Delta_{j\mid0}^*,w_{j\mid0}^*)$ where  
$$\Phi_{j\mid0}(\Delta_{j\mid0},w_{j\mid0}) =\Delta_{j\mid0}(\mathbf{X}_0)+w_{j\mid0}\big[(\widetilde Q_j^*-\Err_0)-\Delta_{j\mid0}(\mathbf{X}_0)\big].$$ 
By construction, $\EE(\widetilde Q_j^*\mid \mathbf{X}_j)=Q_j^*(\mathbf{X}_j)$, and hence $\Delta_{j\mid0}^*(\mathbf{X}_0)=\EE(\widetilde Q_j^*-\Err_0\mid \mathbf{X}_0)$. The oracle risk $R_n^*$ characterizes the intrinsic difficulty of the final regression step after removing nuisance-estimation error. For instance, if $\Delta^*$ is $t$-smooth in $d$ dimensions and $\mathcal L_n$ is minimax optimal, then typically $R_n^* = O_p(n^{-t/(2t+d)})$~\citep{Stone1982}.

The key property underlying the analysis is orthogonality: the stage-wise doubly robust pseudo-outcomes are conditionally unbiased for their target contrasts whenever either the corresponding propensity model or the corresponding auxiliary contrast model is correctly specified.

\begin{prop}[Double robustness of the stage-wise pseudo-outcomes]
\label{prop:dr_property}
Fix an outer fold $k$ and treat the out-of-fold nuisance estimators as fixed. Under Assumptions~\ref{assump:id} and \ref{assump:positivity}, for distinct $j,j^c\in\{1,2\}$, the second-stage pseudo-outcome $\Phi_{j^c\mid j}$ in \eqref{eq:phi_lj_o} satisfies
\[
\EE\!\left[ \Phi_{j^c\mid j}(\Delta_{j^c\mid j}, w_{j^c\mid j})  \mid \mathbf{X}_j,\; S_1=j\right]
=
\Delta^*_{j^c\mid j}(\mathbf{X}_j),
\]
whenever either $\Delta_{j^c\mid j}=\Delta^*_{j^c\mid j}$ or $ w_{j^c\mid j} =  w_{j^c\mid j}^*$. For the first stage,
\[
\EE\!\left[ \Phi_{j\mid0}(\Delta_{j\mid0},w_{j\mid0})  \mid \mathbf{X}_0\right]
=
\Delta^*_{j\mid0}(\mathbf{X}_0),
\]
whenever either $\Delta_{j\mid0}=\Delta_{j\mid0}^*$ or $w_{j\mid0}=w_{j\mid0}^*$.
\end{prop}

\begin{remark}
The second-stage identity is directly feasible because $T_{j^c\mid j}=\Err_{12}-\Err_j$ is defined on the ordered path $(S_1,S_2)=(j,j^c)$. At the first stage, however, the oracle label $T_{j\mid0}=\widetilde Q_j^*-\Err_0$ depends on the latent continuation value $\widetilde Q_j^*$ and is not directly observed. In the feasible procedure, $T_{j\mid0}$ is replaced by $\widetilde Q_j^{(-k)}-\Err_0$, which introduces an additional downstream approximation error. Proposition~\ref{prop:upq-error} shows that this error is controlled by the second-stage contrast estimation error and therefore enters the first-stage oracle inequality in Theorem~\ref{thm:oracle-stage1}.
\end{remark}

\subsection{Asymptotic properties of stage-wise doubly robust estimators}
\label{subsec:asymptotics_stage_wise}

\begin{theorem}[Oracle inequality for second-stage doubly robust estimators]
\label{thm:oracle-stage2}
For $j\in\{1,2\}$, let $\widehat{\Delta}_{\DR,j^c\mid j}$ be the estimator defined in \eqref{eq:delta_dr_lj}, and let $R_{n,j^c\mid j}^*$ be its corresponding oracle risk (Definition 1). Under Assumptions~\ref{assump:id}--5, and whenever the pseudo-outcome consistency precondition for stability holds such that $\|\widehat{\Phi}_{j^c\mid j}-\Phi_{j^c\mid j}^*\|_{L_2(P)}=o_p(1)$, the following oracle inequality is satisfied:
\[
\|\widehat{\Delta}_{\DR,j^c\mid j}-\Delta_{j^c\mid j}^*\|_{L_2(P)}
\le
C_{\mathcal{L}}\,r_\pi r_\Delta
+
R_{n,j^c\mid j}^*
+
o_p(R_{n,j^c\mid j}^*),
\]
where $C_{\mathcal{L}}$ is the boundedness constant of the learner from Assumption 5.
\end{theorem}

Theorem~\ref{thm:oracle-stage2} highlights the doubly robust structure: the estimator remains consistent if either the propensity model or the auxiliary contrast model is consistently estimated. Moreover, under the condition $r_\pi r_\Delta=o_p(n^{-1/2})$, the nuisance contribution becomes asymptotically negligible relative to the oracle risk.

\begin{prop}[Bounding the approximation error]
\label{prop:upq-error}
For $j\in\{1,2\}$, let $\widetilde{Q}_j$ be the estimated continuation value defined in \eqref{eq:Qtilde}, where we suppress the fold index $(-k)$ to treat it as a generic estimator, and let $\widetilde{Q}_j^*$ be the oracle value in \eqref{eq:oracle_upq}. Then
\[
\|\widetilde{Q}_j - \widetilde{Q}_j^*\|_{L_2(P)} \le \|\widehat{\Delta}_{\DR,j^c\mid j} - \Delta_{j^c\mid j}^*\|_{L_2(P)}.
\]
\end{prop}

\begin{theorem}[Oracle inequality for first-stage doubly robust estimators]
\label{thm:oracle-stage1}
For $j\in\{1,2\}$, let $\widehat{\Delta}_{\DR,j\mid0}$ be the estimator defined in \eqref{eq:delta_dr_j0}, and let $R_{n,j\mid0}^*$ be its corresponding oracle risk (Definition 1). Under Assumptions~\ref{assump:id}--5, and whenever $\|\widehat{\Phi}_{j\mid0}-\Phi_{j\mid0}^*\|_{L_2(P)}=o_p(1)$, there exists a constant $C_w$ such that
\[
\|\widehat{\Delta}_{\DR,j\mid0}-\Delta_{j\mid0}^*\|_{L_2(P)}
\le
C_{\mathcal{L}}
\left(
r_{\pi}\,r_{\Delta}
+
C_w\|\widehat{\Delta}_{\DR,j^c\mid j}-\Delta_{j^c\mid j}^*\|_{L_2(P)}
\right)
+
R_{n,j\mid0}^*
+
o_p(R_{n,j\mid0}^*).
\]
\end{theorem}

Theorem~\ref{thm:oracle-stage1} is analogous to the second-stage result, but with one additional term reflecting the sequential nature of the problem: first-stage accuracy depends on the quality of the downstream second-stage estimator used to construct the continuation value.

\subsection{From stage-wise doubly robust learning to COST-Q policy}
\label{subsec:policy_performance}

We evaluate the estimated policy $\widehat{\mathbf{d}}$ via the \emph{Expected Regret}, defined as $\mathfrak R(\widehat{\mathbf{d}}):=\mathbb{E}[Q_0^{\widehat{\mathbf{d}}}(X_0)-Q_0^{\mathbf{d}^*}(X_0)]\ge 0$. The next result links this quantity to the stage-wise $L_2$ estimation errors of the contrast functions.

\begin{theorem}[Upper bound on the expected regret of COST--Q]
\label{thm:policy_regret}
Under Assumptions~\ref{assump:id}--5, let $r_{n,j^c\mid j}:=\|\widehat{\Delta}_{\DR,j^c\mid j}-\Delta_{j^c\mid j}^*\|_{L_2(P)}$ and let $r_{n,j\mid0}:=\|\widehat{\Delta}_{\DR,j\mid0}-\Delta_{j\mid0}^*\|_{L_2(P)}$. Then
\[
\mathfrak R(\widehat{\mathbf{d}})
= O_p\!\Big(r_{n,1\mid0}+r_{n,2\mid0}+r_{n,2\mid1}+r_{n,1\mid2}\Big).
\]
\end{theorem}

To refine this from expected regret, which is linear in the estimation error, to the probability of making a suboptimal decision, we invoke the margin conditions in Assumptions 6 and 7 of Supplementary S.2.

\begin{corollary}
\label{thm:margin_rates_single}
For $j\in\{1,2\}$, the second-stage decision rule $d_j^*(\cdot)$ is governed by the scalar contrast $\Delta_{j^c\mid j}^*(\cdot)$. Under Assumptions~\ref{assump:id}--5 and Assumption 6 applied to $\Delta_{j^c\mid j}^*(\mathbf{X}_j)$, the following hold:
\begin{enumerate}
    \item \textbf{Stage-wise regret bound:}
    \[
    \EE\!\left[\,|\Delta_{j^c\mid j}^*(\mathbf{X}_j)|\,\I\!\left(\widehat d_j(\mathbf{X}_j)\neq d_j^*(\mathbf{X}_j)\right)\right]
    \le \|\widehat{\Delta}_{\DR,j^c\mid j}-\Delta_{j^c\mid j}^*\|_{L_2(P)}.
    \]

    \item \textbf{Misclassification probability:}
    \[
    \pr\!\left(\widehat d_j(\mathbf{X}_j)\neq d_j^*(\mathbf{X}_j)\right)
    \lesssim
    \|\widehat{\Delta}_{\DR,j^c\mid j}-\Delta_{j^c\mid j}^*\|_{L_2(P)}^{\frac{\kappa}{\kappa+1}}.
    \]
\end{enumerate}
\end{corollary}

At the first stage, the decision $d_0$ is three-way rather than binary. An error can therefore arise either from misidentifying whether an additional test is beneficial or from incorrectly ranking the two beneficial test options when both improve upon stopping.

\begin{corollary}[Convergence rates]
\label{coro:margin_costq}
Suppose Assumptions~6 and 7 hold with exponent $\kappa$. Let $r_{n,\cdot}$ denote the stage-wise $L_2$ estimation errors in Theorems~\ref{thm:oracle-stage2} and \ref{thm:oracle-stage1}. Then, For $j\in\{1,2\}$,
\[
\pr\!\left(\widehat d_j(\mathbf{X}_j)\neq d_j^*(\mathbf{X}_j)\right)
\lesssim
r_{n,j^c\mid j}^{\frac{\kappa}{\kappa+1}},
\]
and
\[
\pr\!\left(\widehat d_0(\mathbf{X}_0)\neq d_0^*(\mathbf{X}_0)\right)
\lesssim
r_{n,1\mid0}^{\frac{\kappa}{\kappa+1}} + r_{n,2\mid0}^{\frac{\kappa}{\kappa+1}}.
\]
Consequently, if all stage-wise doubly robust estimators achieve the rate $O_p(n^{-\rho})$, then the policy misclassification probability at each stage is $O_p\big(n^{-\rho\kappa/(\kappa+1)}\big)$, while the total expected regret $\mathfrak R(\widehat{\mathbf{d}})$ decays at the linear rate $O_p(n^{-\rho})$.
\end{corollary}

\begin{remark}
Supplementary~S.4 provides a more granular analysis of \emph{pathwise conditional consistency}, showing that correct decisions at early stages guarantee the optimal convergence of subsequent rules on the realized path.
\end{remark}

\begin{remark}[Estimated value of the learned policy]
In addition to estimating the optimal decision rules, the COST--Q framework naturally yields an internal estimate of the value of the learned sequential testing policy. This is particularly useful in retrospective clinical settings where external fully observed validation data may not be available. Let $\mathcal V(\mathbf d):=\mathbb E\{Q_0^{\mathbf d}(\mathbf X_0)\}$,
so that smaller values correspond to lower expected cost-augmented loss.

To evaluate the learned policy $\widehat{\mathbf d}$, we reuse the cross-fitted pseudo-outcomes constructed in the policy learning procedure. For each outer fold $k$, let $\widehat d_0^{(-k)}$ denote the first-stage decision rule trained on $\mathcal D\setminus \mathcal D_k$, and let $\widehat d_1^{(-k)}, \widehat d_2^{(-k)}$ denote the corresponding second-stage rules. For each subject $i \in \mathcal D_k$, define
\[
\widehat Q_0^{\,i,(-k)}(\widehat{\mathbf d})
=
{\Err}_0^i
+
\widehat{\Phi}_{1\mid0}^i\,\mathbb I\!\left\{\widehat d_0^{(-k)}(\mathbf X_0^i)=1\right\}
+
\widehat{\Phi}_{2\mid0}^i\,\mathbb I\!\left\{\widehat d_0^{(-k)}(\mathbf X_0^i)=2\right\},
\]
where $\widehat{\Phi}_{j\mid0}^i$ is computed using the corresponding fold-specific continuation rule. We then estimate the value of the learned policy by
\[
\widehat{\mathcal V}(\widehat{\mathbf d})
=
\frac{1}{n}\sum_{k=1}^K\sum_{i\in\mathcal D_k}
\widehat Q_0^{\,i,(-k)}(\widehat{\mathbf d}).
\]

This provides a fully data-adaptive estimate of the expected cost-augmented loss under the learned sequential testing policy, and can be computed directly as a by-product of the COST--Q procedure without requiring any external complete-data validation set. In practice, smaller values of $\widehat{\mathcal V}(\widehat{\mathbf d})$ indicate better cost-adjusted performance.
\end{remark}

\section{Simulation Study}
\label{sec:simulation}

We conducted simulation studies to evaluate the finite-sample performance of COST-Q and to compare it with several benchmark methods under both correct specification and nuisance-model misspecification. In both scenarios, the outcome $Y$ is binary, and the goal is to learn a cost-sensitive sequential biomarker acquisition policy. 

We consider two data-generating settings. In Scenario 1, the baseline variable follows $X_0 \sim \mathcal N(0,1)$ truncated to $[-2,2]$. Conditional on $X_0$, the biomarkers satisfy $X_1 \sim \mathcal N(X_0,0.9^2)$ and $X_2\mid X_0 \sim \mathcal N(X_0,0.9^2)$, with both truncated to $[-2,2]$. The outcome is generated as $Y \sim \mathrm{Bernoulli}(q_1)$, where $$q_1=\mathrm{logistic}\!\left(0.15X_0+1.2X_1+1.2X_2+8\sin(2X_1)\sin(2X_2)/2.2\right).$$ 
The acquisition costs are $(c_1,c_2)=(0.01,0.02)$. In Scenario 2, we generate $X_0,X_1,X_2 \overset{iid}{\sim}\mathrm{Unif}(0,1)$ and $Y \sim \mathrm{Bernoulli}(q_2)$, where 
$$q_2=\mathrm{logistic}\bigl(\eta_2(X_0,X_1,X_2)/3.0\bigr),$$ 
and $\eta_2(\cdot)$ contains nonlinear main effects and interaction terms 
(see Supplementary~S.5.3.2 for the full specification). The acquisition costs are $(c_1,c_2)=(0.004,0.002)$.

In both scenarios, missingness is induced through a two-stage behavior policy. The first-stage decision $S_1\in\{0,1,2\}$ follows a softmax model based on $X_0$, while the second-stage continuation decisions follow logistic models based on the observed biomarker history. Full details of the missingness mechanism are provided in Supplementary~S.5.3.1--S.5.3.3.

We compared COST-Q with three benchmark methods: backward outcome weighted learning (BOWL) \citep{zhao2015new}, a complete-case approach (Only-Complete), and a one-time selection rule (One-Time); detailed simulation and learner configurations are provided in Supplementary~S.5.3.4. To assess robustness to nuisance-model misspecification, we considered three settings within each data-generating scenario: A $(\pi\checkmark,\Delta\checkmark)$, B $(\pi\times,\Delta\checkmark)$, and C $(\pi\checkmark,\Delta\times)$, corresponding to correct specification of both nuisance components, misspecification of the propensity model only, and misspecification of the contrast model only, respectively. Because not all benchmark methods involve both nuisance components, Only-Complete does not use propensity models and therefore yields identical results in Settings A and B, whereas BOWL does not involve the contrast model ($\Delta$) and reflects misspecification through its own stage-wise decision model.

For both scenarios, we considered sample sizes $n\in\{400,800,1600,3200,6400,12800\}$, with $R=50$ repetitions and an independent test set of size $5000$. In the main text, we present the distribution of the average total loss, defined as the sum of predictive loss and acquisition cost, at representative sample sizes \(n=400,3200,12800\); the decomposed results for predictive loss and acquisition cost are reported in Supplementary Figures~S2 and S3.

\begin{figure}[!h]
    \centering
    \includegraphics[width=0.95\textwidth]{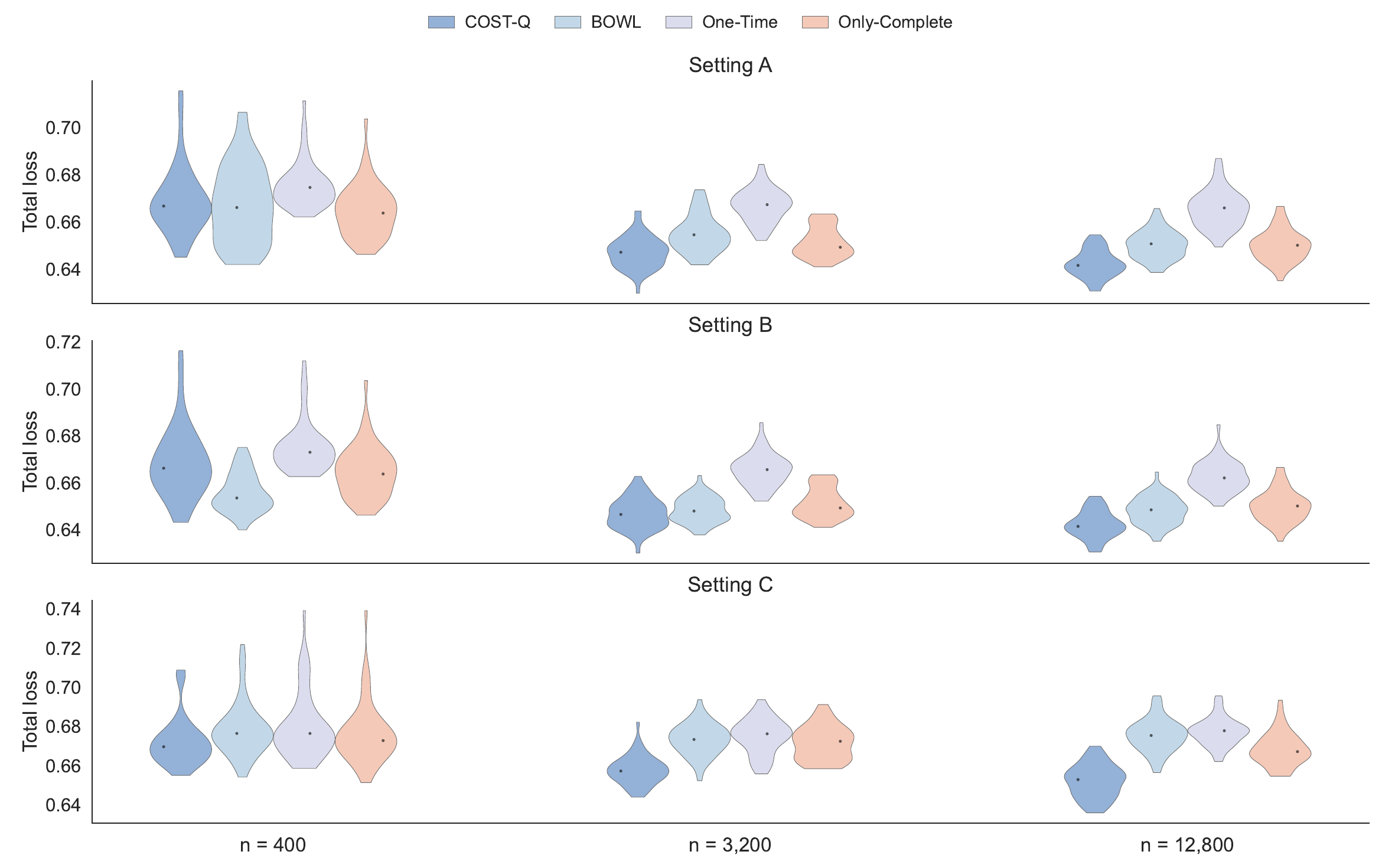}
    \caption{Scenario 1. Violin plots of average total loss across 50 repetitions at representative sample sizes \(n=400,3200,12800\). Rows correspond to settings A/B/C.}
    \label{fig:s1_avg_total_violin}
\end{figure}

\begin{figure}[!h]
    \centering
    \includegraphics[width=0.95\textwidth]{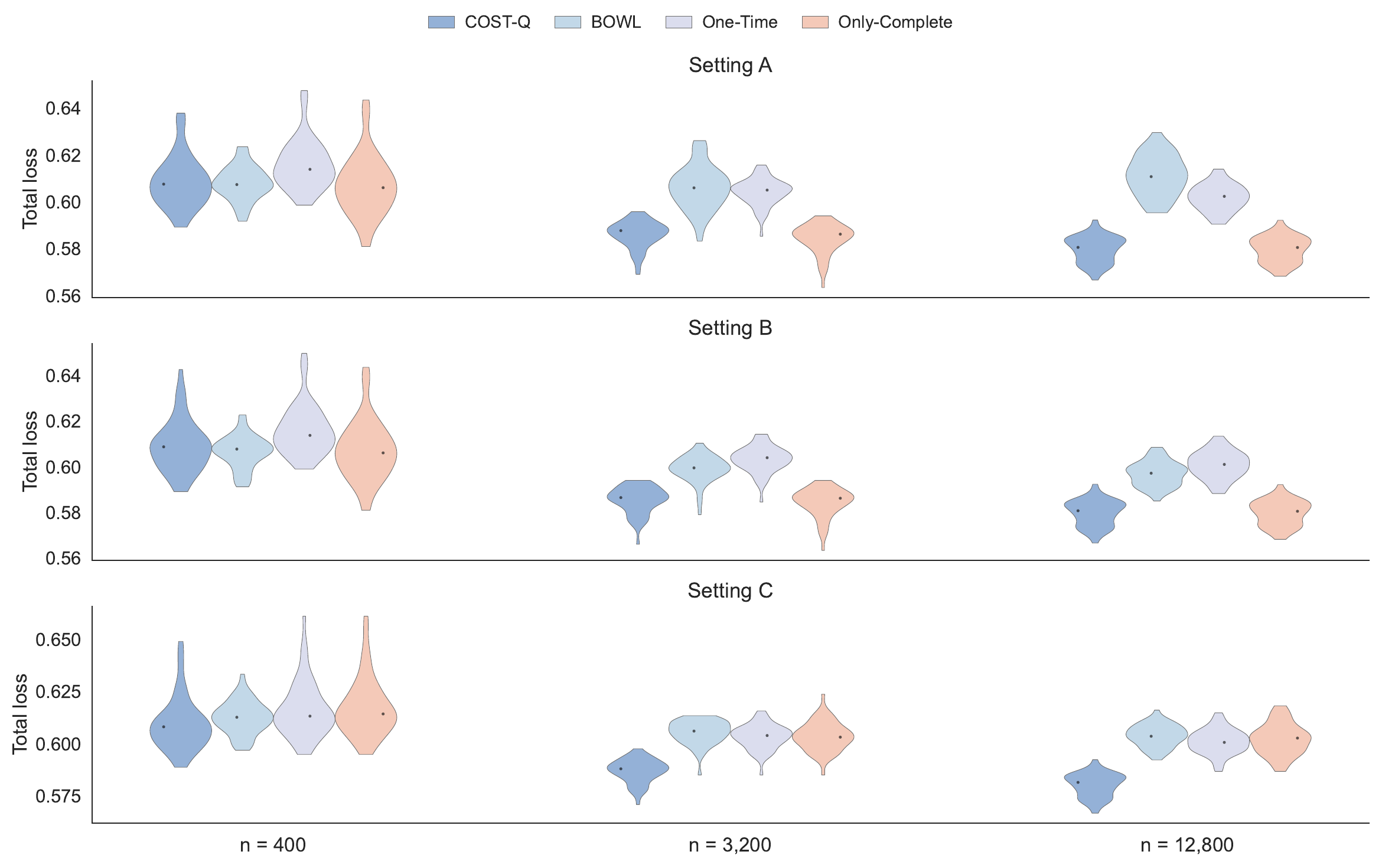}
    \caption{Scenario 2. Violin plots of average total loss across 50 repetitions at representative sample sizes \(n=400,3200,12800\). Rows correspond to settings A/B/C.}
    \label{fig:s2_avg_total_violin}
\end{figure}

Figures~\ref{fig:s1_avg_total_violin} and \ref{fig:s2_avg_total_violin} show that the performance of COST-Q varies across scenario and nuisance setting. In Scenario 1, COST-Q becomes the best-performing method in terms of total loss once the sample size is moderate, with a particularly pronounced advantage under nuisance misspecification, especially in Setting 1(C). In Scenario 2, COST-Q remains competitive across all sample sizes, matching the best-performing benchmark in large samples under Settings 2(A) and 2(B), and exhibiting a clearer advantage in Setting 2(C). Overall, these results highlight the robustness of the doubly robust construction: COST-Q provides the greatest benefit when at least one nuisance component is misspecified, while under well-specified or simpler settings, methods such as Only-Complete can performed comparably.

A closer examination of the loss components (Supplementary Figures~S2 and S3) shows that the advantage of COST-Q is driven primarily by improved predictive performance rather than reduced acquisition cost. In both scenarios, methods that achieve the lowest testing cost often incur substantially higher prediction loss. In contrast, COST-Q consistently attains lower or comparable prediction loss while maintaining a similar, though not always minimal, acquisition cost. This pattern is most evident in Scenario 1 and in Setting 2(C), where gains in prediction accuracy translate into the most favorable overall accuracy-cost tradeoff between accuracy and cost.

\section{Real-world Application}
\label{sec:data}

We evaluate COST-Q using data from the PCA3 Evaluation Trial conducted by the National Cancer Institute Early Detection Research Network (NCI-EDRN) \citep{wei2014can}. 
The study enrolled $743$ participants with elevated PSA or abnormal digital rectal exam (DRE), all of whom underwent diagnostic biopsy. The binary outcome is indicates clinically significant PCa ($Y=1$: high-grade cancer with Gleason score $\geq 7$; $Y=0$: no cancer or low-grade cancer as indicated by a biopsy), with $151$ positive cases. The study aims to assess biomarker-based strategies guiding biopsy decisions, balancing early detection of high grade cancer with avoidance of unnecessary invasive procedures. 

Three diagnostic inputs are considered. The baseline variable $X_0$ is the Prostate Cancer Prevention Trial (PCPT) risk score \citep{thompson2006assessing}, derived from routinely collected clinical variables including serum PSA, age, DRE result, race, and prior history of negative biopsy. The additional bioarmerks are $X_1$, the prostate health index (PHI), a serum-based test combining three PSA isoforms  \citep{sokoll20082}; and $X_2$, the MyProstateScore (MPS), a urine-based biomarker integrating PSA, TMPRSS2-ERG, and PCA3 \citep{tomlins2016urine}. Our goal is to learn a cost-effective individualized sequential testing strategy that accurately identifies high-risk patients for biopsy while minimizing unnecessary testing. A de-identified and slightly perturbed version of the dataset is used for this analysis, where Gaussian noise (10\% of each variable standard deviation) is added to continuous variables (PCPT, PHI, and MPS).

Although all measurements are available in the trial, real-world testing is sequential and incomplete. To reflect this, we impose a two-stage missingness mechanism mimicking clinical practice: first-stage test selection depends on $X_0$, and second-stage continuation depends on observed biomarker history. The same mechanism is applied to all methods to ensure fair comparison. We randomly split the dataset into $70\%$ training ($521$ patients) and $30\%$ testing ($222$ patients). Test costs are assigned proportional to clinical burden, with $c_1=0.012$ for the blood test and $c_2=0.024$ for the advanced urine test. Models are trained on the training set and evaluated on the test set. Performance is assessed using (i) total loss (prediction loss plus test cost), (ii) area under the ROC curve (AUC), and (iii) operating-point metrics, including specificity and geometric mean (G-Mean) at fixed recall levels ($90\%$ and $95\%$) selected on the training set. These thresholds reflect clinical priorities, where missed high-grade cancers are more consequential than false positives. Detailed real-data method and learner configurations are provided in Supplementary~S.5.4.1.

 We also include two fixed reference policies: \textit{always-stop}, which predicts using only baseline information $X_0$, and \textit{always-test-all}, which administers both tests to every patient.
Table~\ref{tab:loss_comparison_A} summarizes the loss components. COST-Q achieves the lowest total loss, while the two fixed reference policies mark the two extremes of testing burden: the always-stop policy attains the lowest testing cost but substantially worsens predictive performance, whereas the always-test-all policy attains the lowest prediction loss but incurs the maximal testing cost. Overall, COST-Q provides a favorable balance between predictive accuracy and testing burden.

\begin{table}[!ht]
\centering
\begin{tabular}{lccc}
\hline
Method & Total Loss & Prediction Loss & Average Cost \\
\hline
COST-Q          & \textbf{0.4212}   & \underline{0.3925} & 0.0287 \\
One-Time        & 0.4637             & 0.4340             & 0.0297 \\
Only-Complete   & 0.4593             & 0.4323             & 0.0270 \\
BOWL            & 0.4489             & 0.4259             & \underline{0.0230} \\
Always-stop     & 0.4704             & 0.4704             & \textbf{0.0000} \\
Always-test-all & \underline{0.4284} & \textbf{0.3924}    & 0.0360 \\
\hline
\end{tabular}
\caption{Loss decomposition on real data (mean over 10 seeds). Bold and underlined indicate the best and second-best values, respectively.}
\label{tab:loss_comparison_A}
\end{table}


To provide a fairer comparison, we additionally performed a matched-cost analysis.  Specifically, for each method we introduced a tuning multiplier $\lambda$ on the test-cost term in the training objective,
\[
\mathcal{L}_{\lambda} = \text{Prediction loss} + \lambda\,\text{Cost},
\]
searched over a grid of $\lambda$ values, and selected the fitted model whose realized average testing cost was closest to a common target budget of $0.027$. We chose $0.027$ because it is attainable for all methods except BOWL and yields a close cost match overall. Unlike the other methods, BOWL adjusts decision boundaries through outcome-weighted classification rather than by directly learning value or contrast functions, so changing $\lambda$ mainly rescales weights and has limited effect on its realized testing frequency. Table~\ref{tab:loss_comparison_budget} shows that, under this approximately matched testing budget, COST-Q achieves the lowest prediction loss and the highest AUC.

\begin{table}[!ht]
\centering
\begin{tabular}{lcccc}
\hline
Method & Total Loss & Prediction Loss & Average Cost & AUC \\
\hline
COST-Q        & \textbf{0.4041}   & \textbf{0.3771}   & \underline{0.0269} & \textbf{0.8181} \\
One-Time      & 0.4224            & 0.3951            & 0.0273             & 0.7936 \\
Only-Complete & \underline{0.4200} & \underline{0.3929} & 0.0271             & \underline{0.8087} \\
BOWL          & 0.4286            & 0.4037            & \textbf{0.0249}    & 0.7895 \\
\hline
\end{tabular}
\caption{Comparison under an approximately matched testing budget of $0.027$. Bold and underlined indicate the best and second-best values, respectively.}
\label{tab:loss_comparison_budget}
\end{table}

Table~\ref{tab:evaluation_comparison_A} reports discrimination and operating-point metrics. COST-Q attains the highest AUC and also achieves the best G-Mean at both recall targets ($90\%$ and $95\%$), as well as the highest specificity at the $95\%$ recall target. The always-test-all policy yields slightly higher specificity at the $90\%$ recall target, but at substantially greater testing cost. Note that always-test-all is a fixed reference policy estimated within the same observed-data sequential framework, rather than an oracle model fitted on fully observed data. Therefore, measuring all tests does not necessarily imply the best discrimination. At the $90\%$ recall level, COST-Q achieves a specificity of $59.5\%$ on the test set, implying that about $59\%$ of men without clinically significant disease who might otherwise be referred for invasive biopsy could potentially avoid the procedure. This represents a substantial improvement over the original trial results, which reported a specificity of approximately $35\%$ at a comparable recall level under full testing with $X_0$ and $X_2$ \citep{sanda2017association}. Overall, these results suggest that COST-Q provides a strong balance among predictive performance, clinically relevant operating-point quality, and testing burden in this real-data analysis. As an auxiliary reference, we also fit a logistic regression model using fully observed $X_0$, $X_1$, and $X_2$ (Supplementary Table~S2). While this model achieves lower total loss, COST-Q achieves higher AUC and better operating-point performance at substantially lower testing cost. This difference reflects the advantage of path-specific decision rules in COST-Q comapred to a single global model.

\begin{table}[!ht]
\centering
\begin{tabular}{lccccc}
\hline
Method & AUC & Spec@90\% & G-Mean@90\% & Spec@95\% & G-Mean@95\% \\
\hline
COST-Q          & \textbf{0.8506}    & \underline{0.5949} & \textbf{0.7170}    & \textbf{0.5186}    & \textbf{0.6937} \\
One-Time        & \underline{0.8328} & 0.5650             & \underline{0.6997} & \underline{0.4915} & \underline{0.6773} \\
Only-Complete   & 0.8074             & 0.4972             & 0.6564             & 0.4011             & 0.6119 \\
BOWL            & 0.7933             & 0.5198             & 0.6537             & 0.4181             & 0.6096 \\
Always-stop     & 0.6330             & 0.2429             & 0.4409             & 0.2090             & 0.4311 \\
Always-test-all & 0.8075             & \textbf{0.6158}    & 0.6921             & 0.4746             & 0.6413 \\
\hline
\end{tabular}
\caption{AUC and operating-point metrics on real data (mean over 10 seeds). Thresholds for recall targets are selected on training data and evaluated on test data. Bold and underlined indicate the best and second-best values, respectively.}
\label{tab:evaluation_comparison_A}
\end{table}

\begin{figure}[!ht]
\centering
\includegraphics[width=0.72\textwidth]{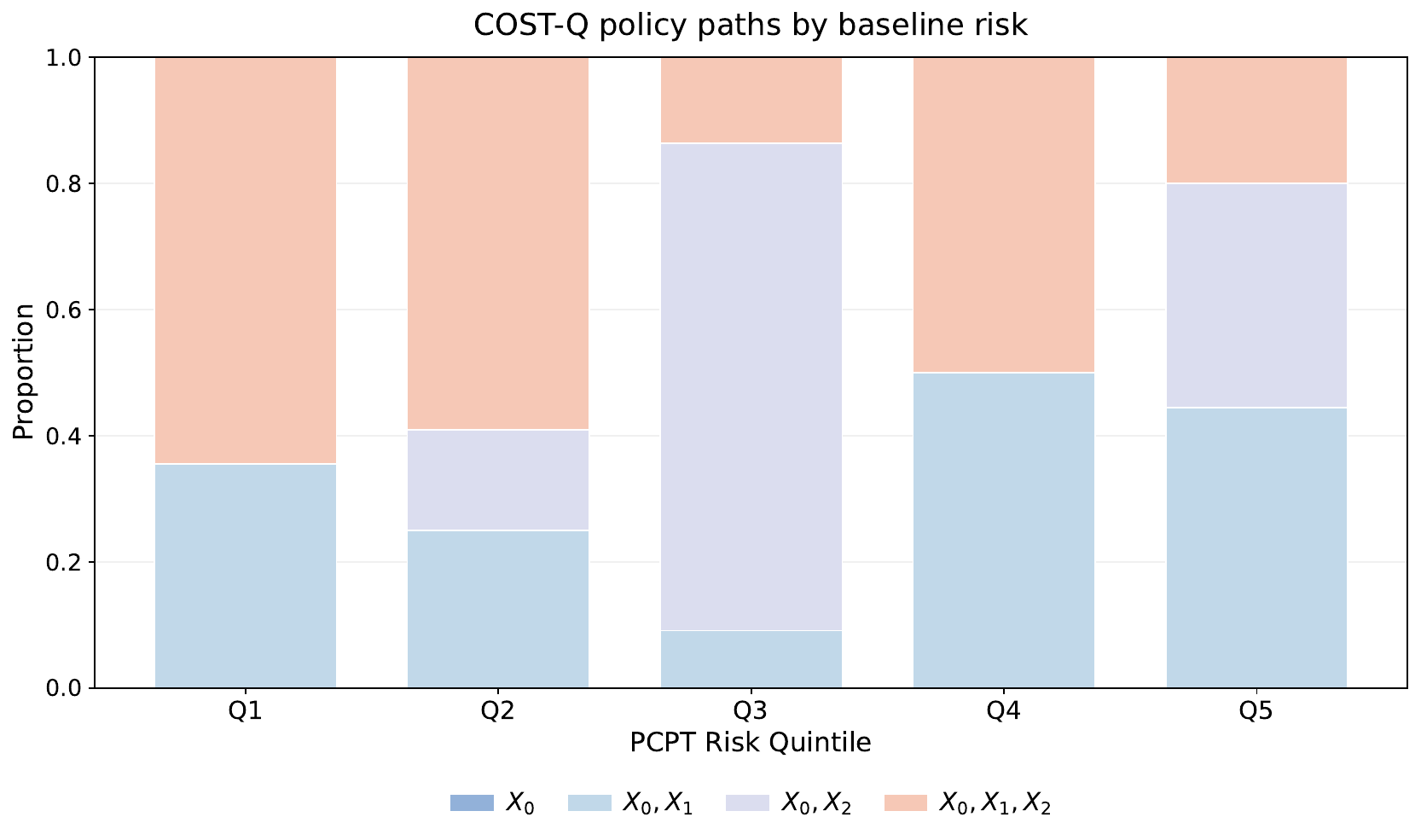}
\caption{Distribution of COST-Q terminal testing paths across quintiles of the baseline PCPT risk score in the test set. Bars show the proportions of subjects who stop at baseline, receive the blood test only, receive the urine test only, or receive both tests.}
\label{fig:costq_quintile_paths}
\end{figure}

To further interpret the learned policies, Table~\ref{tab:policy_behavior_real} summarizes empirical distribution of test paths. COST-Q distributes subjects across multiple-paths, demonstrating adaptive sequential decision-making, unlike the fixed extreme strategies of always-stop and always-test-all. Figure~\ref{fig:costq_quintile_paths} further shows that testing decisions vary across quintiles of baseline PCPT risk, indicating that COST-Q adapts to patient risk profiles rather than applying a uniform strategy. We emphasize that this pattern is descriptive, as the model does not impose monotonicity constraints.

\begin{table}[!ht]
\centering
\begin{tabular}{lccccc}
\hline
Method & Stop at $X_0$ & Take $X_1$ only & Take $X_2$ only & Take both & Avg.\# tests \\
\hline
COST-Q         & 0.000 & 0.329 & 0.257 & 0.414 & 1.414 \\
One-Time       & 0.167 & 0.153 & 0.518 & 0.162 & 0.995 \\
Only-Complete  & 0.126 & 0.090 & 0.532 & 0.252 & 1.126 \\
BOWL           & 0.009 & 0.000 & 0.928 & 0.063 & 1.054 \\
Always-stop    & 1.000 & 0.000 & 0.000 & 0.000 & 0.000 \\
Always-test-all& 0.000 & 0.000 & 0.000 & 1.000 & 2.000 \\
\hline
\end{tabular}
\caption{Empirical policy behavior on the real-data test set. Each entry is the proportion of subjects assigned to the corresponding terminal testing path; Avg.\# tests is the mean number of acquired tests per subject.}
\label{tab:policy_behavior_real}
\end{table}

\section{Discussion and Conclusion} \label{sec:conclusion}

COST-Q provides a general framework for learning cost-sensitive sequential testing policies from observational data with history-dependent informative missingness. The main contribution is the integration of doubly robust estimation with backward Q-learning, which yields consistent stage-wise contrast estimation when either the acquisition model or the contrast model is correctly specified. When both nuisance components are estimated sufficiently well, the resulting estimators attain oracle-type rates and induce policies with controlled regret and misclassification error. The framework is model-agnostic, requiring only a stable final regression learner, and therefore accommodates a wide range of implementations. Across simulations and the prostate cancer application, COST-Q learned adaptive policies that improved the tradeoff between predictive performance and testing burden relative to baselines.

A useful extension is to formulate policy learning under an explicit testing budget rather than through a fixed penalty. Let $R_{\mathrm{pred}}(\mathbf d)$ and $R_{\mathrm{cost}}(\mathbf d)$ denote the expected prediction loss and expected acquisition cost under policy $\mathbf d$. One may instead consider
\[
\min_{\mathbf d}\ R_{\mathrm{pred}}(\mathbf d)
\qquad \text{subject to} \qquad
R_{\mathrm{cost}}(\mathbf d)\le B,
\]
where $B$ is a clinically specified average budget. Our current cost-augmented loss can be interpreted as a Lagrangian relaxation, $R_{\mathrm{pred}}(\mathbf d)+\lambda R_{\mathrm{cost}}(\mathbf d)$, with the present cost scale absorbing the multiplier $\lambda$. Under suitable regularity conditions, varying $\lambda$ traces the same accuracy--cost frontier, although the correspondence need not be exact in finite samples or within restricted policy classes. A budget-constrained version may therefore be especially attractive in practice, since it targets a directly interpretable operating budget and enables matched-budget comparisons across competing methods.

The framework also extends naturally to settings with structural restrictions on the acquisition process. For example, if the reverse path $(S_1,S_2)=(2,1)$ is disallowed, then the nuisance structure simplifies substantially: the reverse-path propensity $\pi_{1\mid 2}^*$ and contrast $\Delta_{1\mid 2}^*$ disappear, leaving only $\pi_{1\mid 0}^*$, $\pi_{2\mid 0}^*$, $\pi_{2\mid 1}^*$ and the contrasts $\Delta_{1\mid 0}^*$, $\Delta_{2\mid 0}^*$, and $\Delta_{2\mid 1}^*$. The same cross-fitted doubly robust Q-learning construction then applies, with the first stage comparing stopping, acquiring $X_1$, and acquiring $X_2$ directly, and the second stage deciding whether to continue from $X_1$ to $X_2$. Because the full-information state is reached through a unique continuation path, the multi-path normalization issue is removed, so the extension is conceptually straightforward and the theoretical guarantees carry over with minor simplification.

Although our simulations use neural networks, COST-Q does not depend on a particular machine-learning architecture. The final regression step for the contrast functions can also be implemented using classical linear smoothers, such as Nadaraya--Watson kernel regression \citep{nadaraya1964estimating, watson1964smooth} or local polynomial regression \citep{fan1996local}. These estimators admit explicit forms and fit naturally within the stability framework used in our theory \citep{buja1989linear, wasserman2006all, kennedy2023ejs}. Detailed formulas for this implementation are provided in Supplementary~S.5.1.

Several limitations remain. As with other inverse-probability-based methods, performance depends on positivity and on sufficiently accurate estimation of the acquisition mechanism in sparsely observed regions of the covariate space. The current formulation also treats test costs as fixed and known, whereas in practice they may vary across patients, institutions, and time. Extending COST-Q to patient-specific costs, larger action spaces, and richer acquisition regimes is an important direction for future work.

Overall, COST-Q offers a principled approach to learning adaptive diagnostic policies from retrospective data in which test acquisition is itself informative. By combining semiparametric identification, doubly robust pseudo-outcomes, and backward dynamic programming, the framework supports individualized testing strategies that can reduce unnecessary measurement while maintaining predictive performance.

{\small

\bibliographystyle{chicago} 
\bibliography{ref}

@article{carroll2016nccn,
	author = {Carroll, Peter R and Parsons, J Kellogg and Andriole, Gerald and Bahnson, Robert R and Castle, Erik P and Catalona, William J and Dahl, Douglas M and Davis, John W and Epstein, Jonathan I and Etzioni, Ruth B and others},
	journal = {Journal of the National Comprehensive Cancer Network},
	number = {5},
	pages = {509--519},
	publisher = {Harborside Press, LLC},
	title = {NCCN guidelines insights: prostate cancer early detection, version 2.2016},
	volume = {14},
	year = {2016}}

@article{papanicolas2018health,
	author = {Papanicolas, Irene and Woskie, Liana R and Jha, Ashish K},
	journal = {JAMA},
	number = {10},
	pages = {1024--1039},
	title = {Health care spending in the United States and other high-income countries},
	volume = {319},
	year = {2018}}

@article{loeb2011complications,
	author = {Loeb, Stacy and Carter, H Ballentine and Berndt, Sonja I and Ricker, Winnie and Schaeffer, Edward M},
	journal = {The Journal of Urology},
	number = {5},
	pages = {1830--1834},
	publisher = {Wolters Kluwer Philadelphia, PA},
	title = {Complications after prostate biopsy: data from SEER-Medicare},
	volume = {186},
	year = {2011}}

@article{blumenthal2024portrait,
	author = {Blumenthal, David and Gumas, Evan D and Shah, Arnav and Gunja, Munira Z and Williams II, Reginald D},
	journal = {New York: the Commonwealth Fund},
	title = {A Portrait of the Failing US Health System},
	year = {2024}}

@article{smart2025inappropriate,
	author = {Smart, Dave and Schreier, Jeff and Singh, Ila R},
	journal = {Archives of Pathology \& Laboratory Medicine},
	number = {3},
	pages = {253--261},
	publisher = {the College of American Pathologists},
	title = {Inappropriate laboratory testing: significant waste quantified by a large-scale year-long study of Medicare and commercial payer reimbursement},
	volume = {149},
	year = {2025}}

@article{anderson2019it,
	author = {Anderson, Gerard F and Hussey, Peter and Varkey, Bianca},
	journal = {Health Affairs},
	number = {1},
	pages = {7--11},
	publisher = {Project HOPE-The People-to-People Health Foundation, Inc.},
	title = {It's still the prices, stupid: Why the US spends so much on health care, and a tribute to Uwe Reinhardt},
	volume = {38},
	year = {2019}}

@article{sanda2017association,
	author = {Sanda, Martin G and Feng, Ziding and Howard, David H and Tomlins, Scott A and Sokoll, Lori J and Chan, Daniel W and Regan, Meredith M and Groskopf, Jack and Chipman, Jonathan and Patil, Dattatraya H and others},
	journal = {JAMA Oncology},
	number = {8},
	pages = {1085--1093},
	title = {Association between combined TMPRSS2: ERG and PCA3 RNA urinary testing and detection of aggressive prostate cancer},
	volume = {3},
	year = {2017}}

@article{thompson2006assessing,
	author = {Thompson, Ian M and Ankerst, Donna Pauler and Chi, Chen and Goodman, Phyllis J and Tangen, Catherine M and Lucia, M Scott and Feng, Ziding and Parnes, Howard L and Coltman Jr, Charles A},
	journal = {Journal of the National Cancer Institute},
	number = {8},
	pages = {529--534},
	publisher = {Oxford University Press},
	title = {Assessing prostate cancer risk: results from the Prostate Cancer Prevention Trial},
	volume = {98},
	year = {2006}}

@article{tomlins2016urine,
	author = {Tomlins, Scott A and Day, John R and Lonigro, Robert J and Hovelson, Daniel H and Siddiqui, Javed and Kunju, L Priya and Dunn, Rodney L and Meyer, Sarah and Hodge, Petrea and Groskopf, Jack and others},
	journal = {European Urology},
	number = {1},
	pages = {45--53},
	publisher = {Elsevier},
	title = {Urine TMPRSS2: ERG plus PCA3 for individualized prostate cancer risk assessment},
	volume = {70},
	year = {2016}}

@article{sokoll20082,
	author = {Sokoll, Lori J and Wang, Yinghui and Feng, Ziding and Kagan, Jacob and Partin, Alan W and Sanda, Martin G and Thompson, Ian M and Chan, Daniel W},
	journal = {The Journal of Urology},
	number = {2},
	pages = {539},
	title = {[-2] proPSA for prostate cancer detection: an NCI Early Detection Research Network validation study},
	volume = {180},
	year = {2008}}

@book{sutton1998introduction,
	author = {Sutton, Richard S and Barto, Andrew G and others},
	publisher = {MIT press Cambridge},
	title = {{Introduction to Reinforcement Learning}},
	volume = {135},
	year = {1998}}

@book{szepesvari2022algorithms,
	author = {Szepesv{\'a}ri, Csaba},
	publisher = {Springer nature},
	title = {{Algorithms for Reinforcement Learning}},
	year = {2022}}

@article{cai2025dynamic,
	author = {Cai, Bryan and Cui, Ying and Fu, Haoda and Lloyd-Jones, Donald M and Zhao, Lihui and Tian, Lu},
	journal = {Journal of the American Statistical Association},
	
	pages = {1--21},
	publisher = {Taylor \& Francis},
	title = {Dynamic Decision Making With Individualized Variable Selection},
	year = {2025}}

@article{wei2014can,
	author = {Wei, John T and Feng, Ziding and Partin, Alan W and Brown, Elissa and Thompson, Ian and Sokoll, Lori and Chan, Daniel W and Lotan, Yair and Kibel, Adam S and Busby, J Erik and others},
	journal = {Journal of Clinical Oncology},
	number = {36},
	pages = {4066--4072},
	publisher = {American Society of Clinical Oncology},
	title = {Can urinary PCA3 supplement PSA in the early detection of prostate cancer?},
	volume = {32},
	year = {2014}}

@article{murphy2003optimal,
	author = {Murphy, Susan A},
	journal = {Journal of the Royal Statistical Society Series B: Statistical Methodology},
	number = {2},
	pages = {331--355},
	publisher = {Oxford University Press},
	title = {Optimal dynamic treatment regimes},
	volume = {65},
	year = {2003}}

@article{tibshirani1996regression,
	author = {Tibshirani, Robert},
	journal = {Journal of the Royal Statistical Society Series B: Statistical Methodology},
	number = {1},
	pages = {267--288},
	publisher = {Oxford University Press},
	title = {Regression shrinkage and selection via the lasso},
	volume = {58},
	year = {1996}}

@article{zhao2015new,
	author = {Zhao, Ying-Qi and Zeng, Donglin and Laber, Eric B and Kosorok, Michael R},
	journal = {Journal of the American Statistical Association},
	number = {510},
	pages = {583--598},
	publisher = {Taylor \& Francis},
	title = {New statistical learning methods for estimating optimal dynamic treatment regimes},
	volume = {110},
	year = {2015}}

@article{ye2024stage,
	author = {Ye, Hanwen and Zhou, Wenzhuo and Zhu, Ruoqing and Qu, Annie},
	journal = {Journal of Machine Learning Research},
	number = {408},
	pages = {1--51},
	title = {Stage-aware learning for dynamic treatments},
	volume = {25},
	year = {2024}}

@article{nadaraya1964estimating,
	author = {Nadaraya, E. A.},
	doi = {10.1137/1109020},
	eprint = {https://doi.org/10.1137/1109020},
	journal = {Theory of Probability \& Its Applications},
	number = {1},
	pages = {141-142},
	title = {On Estimating Regression},
	url = {https://doi.org/10.1137/1109020},
	volume = {9},
	year = {1964}}

@article{watson1964smooth,
	author = {Geoffrey S. Watson},
	issn = {0581572X},
	journal = {Sankhy{\=a}: The Indian Journal of Statistics, Series A (1961-2002)},
	number = {4},
	pages = {359--372},
	publisher = {Springer},
	title = {Smooth Regression Analysis},
	url = {http://www.jstor.org/stable/25049340},
	urldate = {2025-05-08},
	volume = {26},
	year = {1964}}

@article{lloyd2019use,
	author = {Lloyd-Jones, Donald M and Braun, Lynne T and Ndumele, Chiadi E and Smith, Sidney C and Sperling, Laurence S and Virani, Salim S and Blumenthal, Roger S},
	journal = {Journal of the American College of Cardiology},
	number = {24},
	pages = {3153--3167},
	publisher = {American College of Cardiology Foundation Washington, DC},
	title = {Use of risk assessment tools to guide decision-making in the primary prevention of atherosclerotic cardiovascular disease: a special report from the American Heart Association and American College of Cardiology},
	volume = {73},
	year = {2019}}

@article{grundy20192018,
	author = {Grundy, Scott M and Stone, Neil J and Bailey, Alison L and Beam, Craig and Birtcher, Kim K and Blumenthal, Roger S and Braun, Lynne T and De Ferranti, Sarah and Faiella-Tommasino, Joseph and Forman, Daniel E and others},
	journal = {Journal of the American College of Cardiology},
	number = {24},
	pages = {e285--e350},
	publisher = {American College of Cardiology Foundation Washington DC},
	title = {2018 AHA/ACC/AACVPR/AAPA/ABC/ACPM/ADA/AGS/APhA/ASPC/NLA/PCNA Guideline on the Management of Blood Cholesterol: A Report of the American College of Cardiology/American Heart Association Task Force on Clinical Practice Guidelines},
	volume = {73},
	year = {2019}}

@article{weng2017can,
	author = {Weng, Stephen F and Reps, Jenna and Kai, Joe and Garibaldi, Jonathan M and Qureshi, Nadeem},
	journal = {Plos One},
	number = {4},
	pages = {e0174944},
	publisher = {Public Library of Science},
	title = {Can machine-learning improve cardiovascular risk prediction using routine clinical data?},
	volume = {12},
	year = {2017}}

@book{breiman2017classification,
	author = {Breiman, Leo and Friedman, Jerome and Olshen, Richard A and Stone, Charles J},
	publisher = {Chapman and Hall/CRC},
	title = {{Classification and Regression Trees}},
	year = {2017}}

@inproceedings{xu2012greedy,
	address = {Madison, WI, USA},
	author = {Xu, Zhixiang and Weinberger, Kilian Q. and Chapelle, Olivier},
	booktitle = {Proceedings of the 29th International Coference on International Conference on Machine Learning},
	isbn = {9781450312851},
	location = {Edinburgh, Scotland},
	numpages = {8},
	pages = {1299--1306},
	publisher = {Omnipress},
	series = {ICML'12},
	title = {The greedy miser: learning under test-time budgets},
	year = {2012}}

@article{kachuee2018dynamic,
	author = {Kachuee, Mohammad and Darabi, Sajad and Moatamed, Babak and Sarrafzadeh, Majid},
	journal = {IEEE Transactions on Neural Networks and Learning Systems},
	number = {8},
	pages = {2252--2262},
	publisher = {IEEE},
	title = {Dynamic feature acquisition using denoising autoencoders},
	volume = {30},
	year = {2018}}

@inproceedings{huang2018active,
	author = {Huang, Sheng-Jun and Xu, Miao and Xie, Ming-Kun and Sugiyama, Masashi and Niu, Gang and Chen, Songcan},
	booktitle = {Proceedings of the 24th ACM SIGKDD International Conference on Knowledge Discovery \& Data Mining},
	pages = {1571--1579},
	title = {Active feature acquisition with supervised matrix completion},
	year = {2018}}

@article{zhang2013robust,
	author = {Zhang, Baqun and Tsiatis, Anastasios A and Laber, Eric B and Davidian, Marie},
	journal = {Biometrika},
	number = {3},
	pages = {10--1093},
	title = {Robust estimation of optimal dynamic treatment regimes for sequential treatment decisions},
	volume = {100},
	year = {2013}}

@article{ChernozhukovDML2018,
	author = {Chernozhukov, Victor and Chetverikov, Denis and Demirer, Mert and Duflo, Esther and Hansen, Christian and Newey, Whitney and Robins, James},
	doi = {10.1111/ectj.12097},
	eprint = {https://academic.oup.com/ectj/article-pdf/21/1/C1/27684918/ectj00c1.pdf},
	issn = {1368-4221},
	journal = {The Econometrics Journal},
	number = {1},
	pages = {C1-C68},
	title = {Double/debiased machine learning for treatment and structural parameters},
	url = {https://doi.org/10.1111/ectj.12097},
	volume = {21},
	year = {2018}}

@article{kennedy2023ejs,
	author = {Edward H. Kennedy},
	doi = {10.1214/23-EJS2157},
	journal = {Electronic Journal of Statistics},
	number = {2},
	pages = {3008 -- 3049},
	publisher = {Institute of Mathematical Statistics and Bernoulli Society},
	title = {{Towards optimal doubly robust estimation of heterogeneous causal effects}},
	url = {https://doi.org/10.1214/23-EJS2157},
	volume = {17},
	year = {2023}}

@article{robins2000msm,
	author = {Robins, James M. and Hern{\'a}n, Miguel A. and Brumback, Babette},
	doi = {10.1097/00001648-200009000-00011},
	journal = {Epidemiology},
	
	number = {5},
	pages = {550--560},
	title = {Marginal Structural Models and Causal Inference in Epidemiology},
	url = {https://doi.org/10.1097/00001648-200009000-00011},
	volume = {11},
	year = {2000}}

@article{bang2005dr,
	author = {Bang, Heejung and Robins, James M.},
	doi = {10.1111/j.1541-0420.2005.00377.x},
	journal = {Biometrics},
	
	number = {4},
	pages = {962--973},
	title = {Doubly Robust Estimation in Missing Data and Causal Inference Models},
	url = {https://doi.org/10.1111/j.1541-0420.2005.00377.x},
	volume = {61},
	year = {2005}}

@article{Stone1982,
	author = {Charles J. Stone},
	doi = {10.1214/aos/1176345969},
	journal = {The Annals of Statistics},
	number = {4},
	pages = {1040 -- 1053},
	publisher = {Institute of Mathematical Statistics},
	title = {{Optimal global rates of convergence for nonparametric regression}},
	url = {https://doi.org/10.1214/aos/1176345969},
	volume = {10},
	year = {1982}}

@book{fan1996local,
	address = {Boca Raton, FL},
	author = {Fan, Jianqing and Gijbels, Irene},
	publisher = {Chapman \& Hall/CRC},
	title = {{Local Polynomial Modelling and Its Applications}},
	year = {1996}}

@article{buja1989linear,
	author = {Buja, Andreas and Hastie, Trevor and Tibshirani, Robert},
	doi = {10.1214/aos/1176347115},
	journal = {The Annals of Statistics},
	number = {2},
	pages = {453--510},
	title = {Linear Smoothers and Additive Models},
	volume = {17},
	year = {1989}}

@book{wasserman2006all,
	address = {New York, NY},
	author = {Wasserman, Larry},
	doi = {10.1007/0-387-30623-4},
	publisher = {Springer},
	title = {{All of Nonparametric Statistics}},
	year = {2006}}

}

\end{document}